\documentclass[sn-mathphys,Numbered]{sn-jnl}% Math and Physical Sciences Reference Style
\usepackage{xcolor, soul}
\usepackage{graphicx}%
\usepackage{multirow}%
\usepackage{amsmath,amssymb,amsfonts}%
\usepackage{amsthm}%
\usepackage{mathrsfs}%
\usepackage[title]{appendix}%
\usepackage{xcolor}%
\usepackage[most]{tcolorbox}
\newtcolorbox{highlighted}{colback=yellow,coltext=red,breakable}
\usepackage{textcomp}%
\usepackage{manyfoot}%
\usepackage{booktabs}%
\usepackage{algorithm}%
\usepackage{algorithmicx}%
\usepackage{algpseudocode}%
\usepackage{listings}%
\usepackage{subcaption}
\usepackage{float}
\usepackage{times}
\usepackage{latexsym}
\usepackage{arabtex}
\usepackage{utf8}
\setcode{utf8}
\usepackage{makecell}
\usepackage{comment}
\usepackage{graphicx}
\usepackage[T1]{fontenc}
\usepackage[utf8]{inputenc}
\usepackage{microtype}
\usepackage{inconsolata}
\usepackage{arabtex}  

\begin{document}

\title[Detecting Suicidality]{Detecting Suicidality in Arabic Tweets Using Machine Learning and Deep Learning Techniques}

%%=============================================================%%
%% Prefix	-> \pfx{Dr}
%% GivenName	-> \fnm{Joergen W.}
%% Particle	-> \spfx{van der} -> surname prefix
%% FamilyName	-> \sur{Ploeg}
%% Suffix	-> \sfx{IV}
%% NatureName	-> \tanm{Poet Laureate} -> Title after name
%% Degrees	-> \dgr{MSc, PhD}
%% \author*[1,2]{\pfx{Dr} \fnm{Joergen W.} \spfx{van der} \sur{Ploeg} \sfx{IV} \tanm{Poet Laureate} 
%%                 \dgr{MSc, PhD}}\email{iauthor@gmail.com}
%%=============================================================%%

\author*[1]{\fnm{Asma} \sur{Abdulsalam}}\email{aabdulsalam0012@stu.kau.edu.sa}

\author[1]{\fnm{Areej} \sur{Alhothali}}\email{aalhothali@kau.edu.sa}
%\equalcont{These authors contributed equally to this work.}

\author[2]{\fnm{Saleh} \sur{Al-Ghamdi}}\email{syalghamdi@kau.edu.sa}
%\equalcont{These authors contributed equally to this work.}

\affil*[1]{\orgdiv{Department of Computer Science, Faculty of Computing and Information Technology}, \orgname{King AbdulAziz University}, \orgaddress{\city{Jeddah}, \postcode{21589}, \country{Saudi Arabia}}}

\affil[2]{\orgdiv{Department of Psychology, Faculty of Educational Graduate Studies}, \orgname{King AbdulAziz University}, \orgaddress{\city{Jeddah}, \postcode{21589}, \country{Saudi Arabia}}}

%\affil[3]{\orgdiv{Department}, \orgname{Organization}, \orgaddress{\street{Street}, \city{City}, \postcode{610101}, \state{State}, \country{Country}}}

%%==================================%%
%% sample for unstructured abstract %%
%%==================================%%
\abstract{
Social media platforms have revolutionized traditional communication techniques by enabling people globally to connect instantaneously, openly, and frequently. People use social media to share personal stories and express their opinion. Negative emotions such as thoughts of death, self-harm, and hardship are commonly expressed on social media, particularly among younger generations. As a result, using social media to detect suicidal thoughts will help provide proper intervention that will ultimately deter others from self-harm and committing suicide and stop the spread of suicidal ideation on social media. To investigate the ability to detect suicidal thoughts in Arabic tweets automatically, we developed a novel Arabic suicidal tweets dataset, examined several machine learning models, including Naïve Bayes, Support Vector Machine, K-Nearest Neighbor, Random Forest, and XGBoost, trained on word frequency and word embedding features, and investigated the ability of pre-trained deep learning models, AraBert, AraELECTRA, and AraGPT2, to identify suicidal thoughts in Arabic tweets. The results indicate that SVM and RF models trained on character n-gram features provided the best performance in the machine learning models, with 86\% accuracy and an F1 score of 79\%. The results of the deep learning models show that AraBert model outperforms other machine and deep learning models, achieving an accuracy of 91\% and an F1-score of 88\%, which significantly improves the detection of suicidal ideation in the Arabic tweets dataset. To the best of our knowledge, this is the first study to develop an Arabic suicidality detection dataset from Twitter and to use deep-learning approaches in detecting suicidality in Arabic posts.}


%%\pacs[JEL Classification]{D8, H51}

%%\pacs[MSC Classification]{35A01, 65L10, 65L12, 65L20, 65L70}

\maketitle

\section{Introduction}

Approximately $3.96$ billion people actively accessed the internet worldwide~\cite{1astoveza2018suicidal}. Millions of people use social media regularly, such as chat rooms, social networking sites, blogging sites, and social networking platforms. Social media networking sites such as Facebook, Twitter, Snapchat, and other social networking sites allow users to exchange information and interact with others. Twitter is a free social media broadcast site that allows registered users to communicate with others through 280-character messages called "tweets." This popular platform enables users to say or express whatever they want, whether positive or negative. A significant number of users utilize social media networks to convey their feelings, experiences, thoughts, difficulties, and concerns~\cite{15n10.1145/2858036.2858207}. Self-harming thoughts, death, and suicidal ideation are among the most popular topics discussed on social media. The intended attempt to end the person's own life is referred to as suicide~\cite{2nock2008suicide}. 

Suicide is a phenomenon that arises from a complex interaction of social, biological, cultural, psychological, and spiritual variables~\cite{3beck1979assessment}. Suicide is a manifestation of underlying suffering that is brought on by a variety of events, such as underlying mental illnesses that create psychological pain~\cite{4liu2020suicidal}. Suicidal behavior includes three types: suicidal behavior, suicidal planning, and suicidal ideation~\cite{2nock2008suicide,3beck1979assessment,4liu2020suicidal}. Suicidal ideation refers to the thought or purpose of taking one's own life without actually attempting it. A suicide attempt, on the other hand, is a self-harming act that might lead to death with the intended purpose of dying, as opposed to a suicide plan, which is a specific method a person can take to terminate his/her life~\cite{2nock2008suicide,3beck1979assessment,4liu2020suicidal}. Suicide affects individuals, families, communities, and even countries~\cite{4liu2020suicidal}. Suicide is the second leading cause of death among young people internationally. It causes more deaths than diabetes, liver disease, stroke or infection~\cite{5weber2017psychiatric}. Due to the stigma associated with mental disorders, more than 40\% of people who seek primary care are unwilling or hesitant to discuss their depressive symptoms. Suicidal thoughts and actions require immediate attention, and there is no effective method for assessing, managing, and preventing suicide~\cite{5weber2017psychiatric}.

Traditional methods for suicidality risk assessment rely on psychologists' expertise and self-report questionnaires~\cite{4liu2020suicidal}. The Patient Health Questionaire-9 (PHQ-9) and Columbia Suicide Severity Rating Scale (C-SSRS) are two examples of tools used for screening suicide and identifying depressive symptoms~\cite{5weber2017psychiatric}. Despite the quickness and efficiency of these methods, participant concealment makes them vulnerable to misleading negative results. They are also challenging to carry out over an extended period of time or on a massive scale~\cite{5weber2017psychiatric}.
Therefore, the task of detecting suicidality has attracted researchers from a variety of fields to investigate linguistic and psychological indicators as well as other factors that could help diagnose and identify individuals with suicidal thoughts~\cite{4liu2020suicidal}.

%In the present research, we reviewed several studies that used different languages, and found that English was the most frequently used language, while Chinese language was the second most used language, followed by Spanish language, Russian~, Japanese, Filipino or Taglish language were also used, but only one study on Arabic language was performed using a translated English dataset.

Social media posts can offer valuable insights into an individual's emotional and psychological well-being~\cite{15n10.1145/2858036.2858207}. Many individuals are unable to share their personal experiences or express their emotions in person and may instead choose to write about their feelings or suicidal intentions in blogs or social media posts. Unfortunately, these posts are often disregarded or overlooked. However, this information can be useful in conducting large-scale screenings for suicidal behaviors. Studies in the field investigated suicidality content written in various languages, including Chinese~\cite{17huang2014detecting}, Spanish, Russian~\cite{22rajesh2020suicidal}, Japanese~\cite{24ramirez2020detection}, and Filipino or Taglish~\cite{1astoveza2018suicidal}. English was the most frequently used language~\cite{word2ODEA2015183,32n10.1145/2700171.2791023,14vioules2018detection,15moulahi2017dare,16ma2020dual,17huang2014detecting}, followed by Chinese~\cite{19narynov2019comparative,20du2018extracting,25na13010007,word1FAHEY2020112960} and Spanish~\cite{9ji2018supervised,26chiroma2018suiciderelated}. However, there was only one such study using the Arabic language, which used a translated English dataset~\cite{9966481}.

%%%%%7/29
%Social media has been used as a powerful tool to determine the psychological states of its users~\cite{15n10.1145/2858036.2858207}.
A huge number of Twitter posts are written in Arabic, and millions of its users are from Arab countries. A total of 15 million Twitter users are from Saudi Arabia alone~\cite{16nhttps://doi.org/10.48550/arxiv.2004.04315}. Arabic is the predominant language of $422$ million people in over $27$ nations~\cite{17nBOUDAD20182479}. 
%as reported by Statista in 2020~\cite{16nhttps://doi.org/10.48550/arxiv.2004.04315}.
The Arabic language has a large lexicon, and different varieties of the language exist: Classical Arabic, Modern Standard Arabic (MSA), and dialects or colloquial~\cite{17nBOUDAD20182479,18n10.1145/1644879.1644881}. %The language of the Quran refers to it as classical Arabic. 
MSA is primarily used in formal speech and writing, while the dialectical or colloquial often varies from one country to another and is used as the main language of communication~\cite{17nBOUDAD20182479}. Arabic varieties may have some properties in common, but each has its own lexicon, grammar, and morphology~\cite{18n10.1145/1644879.1644881}. The morphology and orthography of Arabic language text are the main difficulties facing NLP researchers~\cite{17nBOUDAD20182479,19n7945623}. Most MSA text lacks orthographic representation for short vowel letters as these letters are represented by diacritical marks, which are not used in MSA. Therefore, the meanings of words in isolation can be ambiguous~\cite{17nBOUDAD20182479,18n10.1145/1644879.1644881}. Morphology plays an important role in Arabic because it is a derivational language and highly structured~\cite{18n10.1145/1644879.1644881}. An Arabic word may differ based on morphological elements such as root, stem, part of speech (POS), and affix~\cite{17nBOUDAD20182479}. Even letters in the Arabic language have different shapes based on their position in a word~\cite{18n10.1145/1644879.1644881}. 
%%%%%%

%%%%%%%%%%%this paper

A review of related works in the field revealed a lack of suicidal research on Arabic social media in general and automatic detection or identification of Arabic suicidal ideation using machine learning in particular. Therefore, developing machine learning techniques to extract linguistic indications of suicidal thoughts in Arabic will help provide an accurate and effective mechanism to conduct an extensive screening through social media platforms with the aim of detecting suicidal ideations and potentially preventing suicidal behaviors. To the best of our knowledge, this is the first study that builds a machine or deep learning model to detect suicidal content in Arabic.
%In this study, a deep learning-based approach was developed to detect suicidal content.

%\hl{
This research proposes an Arabic suicide ideation detection framework to analyze tweets with various features and explore the possibility of monitoring suicidal behaviors. The contribution of this research is three fold. First, we have developed the first Arabic suicidality detection dataset comprising 5,719 Arabic tweets that were collected between 23 August 2021 and 21 April 2022 from different Arabic-speaking countries using different Arabic dialects; 1,429 were labeled as Suicidal tweets, and 4,290 were Non-Suicidal. The labeling process was conducted by two annotators, one of whom holds a Ph.D. in psychology and has been working on several cyber-psychology projects. %To summarize, the contribution of this research is three folds. First, this research developed the first suicidal ideation dataset in Arabic. 
Second, several machine learning models were examined to automatically detect suicidal thoughts with different types of feature representations that include word frequency features (BOW, unigram TF-IDF, n-gram TF-IDF, and character n-gram TF-IDF) and word embedding (i.e., Arabic word2vec and FastText) features. Third, pre-trained deep learning models with transformer-based architecture and attention mechanism (AraBert, AraELECTRA, and AraGPT2) was utilized for the prediction of suicidal thoughts in Arabic tweets.%}

 %more emphasis on the contribution of the paper .. add summary of the contribution... we have developed the first Arabic suicidality detection dataset that consisted of .. the dataset was annotated by physiologist  expert in the field .. a deep learning based approach was developed in this study to detect suicidal content.

    The rest of this paper is organized as follows: Section{~\ref{Sec:related works}} reviews the previous work on detecting suicidality. The methodology used for dataset and feature extraction and the algorithms used in the classification process are discussed in detail in Section~{\ref{Sec:Methodology}}. The results of this study are discussed in Section~{\ref{Sec:result}}. Section~{\ref{Sec:conclusion}} concludes the study and looks ahead to future works.%}

\section{Related Works}
~\label{Sec:related works} 

Today with the large and growing numbers of active users on social media, many users use their social media accounts to share and write on a daily basis. This trend of using social media as a modern-day diary can help reveal and analyze part of users' personalities and mental states. Researchers have investigated various machine learning and deep learning techniques along with various types of features to identify suicidal vs. non-suicidal content or determine suicidal risk severity based on social media users' generated content. The classification approaches used in the field can be divided into traditional machine learning models, such as support vector machine (SVM), Naive Bayes (NB), Random Forest (RF), Logistic regression (LR), Decision tree (DT), Artificial Neural Network (ANN), ensembling techniques such as Voting, Stacking, Boosting, Bagging, XGBoost, AdaBoost, and deep learning model such as Long short-term memory (LSTM), Convolutional Neural Networks (CNNs), Recurrent Neural Networks (RNNs), Gated Recurrent Unit (GRU), and Transformer Networks e.g., Bidirectional Encoder Representations from Transformers (Bert). The features used in the field can be divided into statistical (e.g., length of post, number of words and characters), temporal, linguistic (e.g., emotional and sentiment using Linguistic Inquiry and Word Count (LIWC) dictionary~\cite{doi:10.1177/0261927X09351676}), syntactic (e.g., part of speech tagging information), Word Frequency (e.g., bag of word model (BOW), or  Term Frequency-Inverse Document Frequency (TF-IDF)), word embedding (word2vec~\cite{mikolov2013distributed} and Glove~\cite{pennington2014glove}), and topic features (e.g., Latent Dirichlet allocation (LDA)~\cite{blei2003latent}). This section summarizes the latest research on suicidal detection with the main focus on text classification. 

%%%%%%%%%%%%%%%%%% Machine learing approaches 
%1
Several researchers have used supervised learning machine learning models for suicidal ideation detection.~\citet{word2ODEA2015183} examined the possibility of determining the level of concern from Twitter post content. They developed two text classifiers using machine learning algorithms, which are LR and SVM. The word frequency and the weighting word frequency (i.e., TF-IDF) with filter and TF-IDF with no filter variations of the feature space were used to test these techniques. TF-IDF with filter involves removing frequent terms that appeared more than a certain number of times in the document, while word frequency refers to the original word frequency without weighting, and TF-IDF with no filter refers to the traditional TF-IDF with no word removal. The results showed that SVMs with TFIDF no-filter had the highest accuracy. Another study~\cite{33n8527039} used four machine classifiers which are DT, NB, RF, and SVM, on the same datase~\cite{32n10.1145/2700171.2791023}. The data was divided into two datasets—one set for binary classification (Suicide and Flippant reference to suicide) and another for multi-class classification (Suicide, Flippant, and Non-Suicide) classes. The used features are Part of Speech (POS), Bag of Words (BOW), and Inverse Document Frequency (IDF). The results show that DT gave the highest accuracy for the multi-class classification. 
%DT had the best F1-measure of 0. 879 and 0.790 accuracies for multi-class dataset.
A study by~\citet{26chiroma2018suiciderelated} conducted on the same dataset~\cite{32n10.1145/2700171.2791023} and using the same pre-processing techniques aimed to evaluate the performance of the Prism algorithm compared to standard machine learning algorithms (SVM, DT, NB, and RF) with BOW features. The results indicated that the Prism algorithm outperformed the other classifiers in all performance measures.
%A total of 2,000 tweets or instances were collected, however, after rigorous pre-processing only 1060 instances were left for the experiment.
%7

\citet{26new10.1007/978-3-030-37429-7_17} developed a method to detect suicidal content on the Sina Weibo microblogging platform. They used three different sets of linguistic features, including an automated machine learning dictionary, a Chinese suicide dictionary, and Simplified Chinese Micro-Blog Word Count (SCMBWC). These feature sets were separately applied with the SVM, DT, and LR algorithms to classify the content into one of six classes. The results indicated that the SVM algorithm with the feature set extracted using an automated machine learning dictionary from real blog data with N-gram gave the highest accuracy. % using a logistic regression model. %More than 65,352 messages were crawled from Sina Weibo, only 8,548 blogs were labeled as suicide blogs. Two machine learning algorithms were used (SVM, DT) to build a classification model with three features sets (automated machine learning dictionary, Chinese suicide dictionary, and Simplified Chinese Micro-Blog Word Count(SCMBWC)).  
%Each feature set was used with SVM, DT, and logistic regression algorithms with the three features sets separately to generate six detection results.% Those were input to a logistic regression model. 
%It has been found that SVM with a feature set extracted using an automated machine learning dictionary from real blogs data-driven by N-gram gave the highest accuracy. 
 \citet{15moulahi2017dare} proposed a probabilistic framework based on Conditional Random Fields (CRF) to track suicidal ideation. They used three sets of features which are the syntactic features (POS), linguistic features (i.e., psychological and emotional lexicon features), and contextual features (i.e., posts observed at previous and upcoming sessions). It has been noticed that the CRF model outperformed other machine learning methods, such as SVM, NB, J48, RF, and ANN. 
%%%%%%%%%%%%%%% lexicon based approches
% move it to the lexicon dictionary based methods
\citet{22rajesh2020suicidal} utilized Vader sentiment analysis to assign a sentiment score to each word in their study. They employed various classifiers, including NB, RF, XGBoost, and LR, to distinguish sentences into positive or neutral categories. Several preprocessing techniques were employed, including statistical features, tokenization stemming, BOW, and word frequency, after cleaning the data. They achieved the highest accuracy using the RF method when considering all sets of features. 
 
 %%% ensambling methods

 Ensembling machine learning models were also investigated for suicidal ideation detection.~\citet{32n10.1145/2700171.2791023} incorporated SVM and NB as an ensemble approach known as rotation forest trained on lexical, structural, emotive, and psychological features extracted from Twitter posts to distinguish between more concerning content, such as suicidal ideation content and other suicide-related content, such as suicide reporting, memorial, campaigning, and support. The rotation forest approach was compared to three classifiers, SVM, NB, and J48, and an integrated model of SVM and NB. The experiment showed a better performance when SVM and NB were integrated.  
 %lexical, structural, emotive and psychological features extracted from Twitter posts
 %
 ~\citet{9528252} developed a machine learning model to predict whether or not a tweet has suicide ideation. The dataset used consisted of 9,119 tweets, with 5,121 non-suicidal tweets and 3,998 suicidal tweets. In this research, various machine learning models were used, including SVM, DT, LR, NB, K-NN, and different ensemble models, namely AdaBoost, Gradient Boost, Bagging, CatBoost, XGBoost, and Voting Classifier(VC) were also implemented. The results indicated that the VC achieved the best accuracy. It was used as an estimator with three classifiers, LR, SVM, and DT, that gave the best result of all other combinations.  % with 0.90,0.91, 0.89, 0.90 for accuracy, precision, recall, and F1-score, respectively. A dataset consisting of 3998 suicide-alarming tweets and 5121 non suicidal tweets was used.
A dataset of individuals who expressed suicidal thoughts on Twitter was developed by~\cite{9388638}. The dataset consisted of 1,897 tweets that were gathered using keywords extracted from a previous study~\cite{COLOMBO2016291} and various web forums. The annotation was done by a human annotator and a psychiatric expert. Several machine learning techniques were used, including SVM, LR, MNB, BNB, RF, and DT. Three ensemble learning techniques, including Voting Ensemble and AdaBoost, were also used. The results indicated that LR outperforms other models. % with LR outperforms other models with 0.79 accuracy, 0.48 precision, and 0.12 recall. 

\citet{17huang2014detecting} utilized a real-time system that employed machine learning and a psychological lexicon dictionary to detect suicidal ideation. They analyzed the social media platform Weibo and identified 53 users who had posted suicidal content before their deaths. They employed various classifiers, such as SVM, NB, LR, J48, Rotation Forest (RF), and Sequential Minimal Optimization (SMO), with three N-gram features and a psychological lexicon dictionary. The SVM classifier had the best performance compared to the other classifiers.%, with an F1-measure of 68.3%, a precision of 78.9%, a recall of 60.3%, and an accuracy of over 94%. 
%This positive outcome indicates the possibility of developing a suicide detection system that may assist psychologists in suicidal identification and will be an excellent tool for suicide prevention. 
%Other studies have utilized lexicon or dictionary based features to identify suicidal content.
~\citet{9678419} aimed to improve the machine learning model that uses the grey wolf optimizer (GWO) to identify a person contemplating suicide. GWO was combined with a machine learning algorithm to predict Twitter suicidal users. Tweets that mention depression, self-harm, and anxiety were gathered, and only tweets with signs associated with suicide were collected. A total of 193,720 tweets were collected and annotated. Different machine learning models were implemented that include NB, LR, SVM, and DT. The results show that LR-GWO trained on unigram gave the best accuracy in comparison with the other machine learning models.

%Machine classifiers were built to classify suicide-related text on Twitter~\cite{32n10.1145/2700171.2791023}. Twitter text was classified into suicidal intent or other suicide-related topics e.g., campaigning and support, memorial or reporting of suicide. The dataset was collected from twitter after exacting keywords from four known websites identified as a discussion website for suicidal support and prevention. Four million posts were collected using suicidal keywords, in addition to different tweets gathered using the names of reported suicide cases. Structural, lexical, psychological, and emotive were the most popular classifiers in classifying suicidal conetnt using different machine learning methods such as SVM, DT, and NB. Also, they incorporated SVM and NB as an ensemble approach known as Rotation Forest. They tested rotation forest approach with the three classifiers and another one with only SVM and NB. The experiment showed a better performance when SVM and NB were integrated.
%The dataset was randomly sampled from both datasets with 800 suicidal tweets, and 200 undirect suicidal ideation tweets.
%Therefore, the DT was dropped from the ensemble approach. RF gained 0.690 for F1-measure, precision performance of 0.644, and recall of 0.744.

%%%%%%%%%%%%%%%%%%5
Supervised deep learning has been used by several researchers to identify suicidal content. \citet{9ji2018supervised} used different classifiers from both traditional machine learning models and deep learning models to identify online suicidal users through their online content. The authors used several models such as SVM, RF, gradient boost classification tree (GBDT), XGBoost, multilayer feed-forward neural net (MLFFNN), and long short-term memory (LSTM). Several sets of features were used, including statistical, linguistic, syntactic, topic features, and word embedding. Combining different types of features such as statistical, topic, TF-IDF, POS, and LIWC increased the models’ accuracy. \citet{25na13010007} proposed a model to detect suicidality on the Reddit platform. Two models were combined Long Short-Term Memory (LSTM) and Convolutional Neural Network (CNN) to classify Reddit content in one of multiple classes. TF-IDF, BOW and statistical features were used with four machine learning models (SVM, NB, RF and GXBoosting) and Word2vec was used with two deep learning models (LSTM and CNN). The proposed LSTM-CNN hybrid model outperformed other deep and machine learning algorithms used in this study. %improved the overall performance. % with $93.8\%$, F1-score $93.4\%$, recall $94.1\%$ and Precision $93.2\%$.
%~\citet{36n8269767} studied online users’ textual and behavior features to detect sudden changes in users’ online behavior and identify suicide-related posts. The authors combined both textual and behavioral features and then passed them a martingale framework for detection. They found that the two-step classification performed well in the test set.
%proposed an approach to detecting suicidal thoughts to identify sudden changes in users' online behavior by analyzing users' behavioral and textual features. They collected $5,446$ tweets using special key phrases obtained from a generated list of suicide risk factors and warning signs. Eight researchers and a mental health professional then manually annotated tweets
%Vioules et al. detect the change in the data streams by passing textual and behavior features to a martingale framework. They needed two datasets sufficiently large annotated set and another smaller set of selected Twitter users to study their history. They found that the two-step classification performed well in the test set. They reached $82.9\%$ precision, $81.7\%$ for recall, and F1-score\cite{14vioules2018detection}.
%%%%%new LR

~\citet{info:doi/10.2196/34705} examined the relationship between social media content consumption and suicidal behavior. They created a dataset of 3,202 English tweets and manually classified them into 12 categories, such as personal stories of suicidal ideation, coping and recovery, spreading awareness, prevention information, and irrelevant tweets. They developed several machine learning models to perform multi-class and binary classification tasks. The majority classifier used TF-IDF with a linear SVM, as well as two deep learning models, BERT and XLNet. The first task involved categorizing posts into six key content categories related to suicide prevention, while the second task used binary classification to distinguish between posts in the 12 categories that referred to actual suicide and those in off-topic categories that used terms related to suicide in a different context. Both deep learning models performed similarly in both tasks, outperforming the SVM with TF-IDF except for the suicidal ideation and attempts category. BERT's model achieved the highest overall scores in binary classification.

Around 50,000 tweets were collected by~\citet{technologies10030057} to detect suicidal ideation using machine learning and deep learning classifiers. % The annotation is done in two steps. First, using python tool TextBlob and VADER to annotate tweets by extracting the sentiment polarity (positive, negative, or neutral). Second, review all tweets manually into two classes (suicidal, Non-suicidal). 
Several deep learning models were used, such as LSTM, Bi-directional LSTM (BiLSTM), GRU, Bi-directional GRU (BiGRU), and combined model of CNN and LSTM (C-LSTM). These deep learning models were compared to other traditional machine learning approaches such as RF, SVM, Stochastic Gradient Descent classifier (SGD), LR, and Multinomial Naive Bayes (MNB) classifier. Their results showed that RF model reached the highest classification score among machine learning methods. However, word embedding training improved the performance of DL models, with the BiLSTM model achieving higher precision and F1-score. 
~\citet{9753295} developed a deep learning approach aimed at analyzing online Twitter posts and identifying any characteristics that could indicate the presence of suicidal tendencies. To develop the model, a total of 188,704 tweets were extracted from 1,169 users and manually annotated into two classes: suicide and non-suicidal posts. A variety of features were extracted, including sentiment analysis, emoticons, TF-IDF, statistics, topic-based features, and temporal features. The results showed that LDA with Trigram, TF-IDF, statistics, temporal (time of information publishing on Twitter), Emoticons, and Sentiment Analysis features had the best performance for detecting suicidal ideation in comparison with the other feature combinations.% with an accuracy of 0.87, precision of 0.83, and 0.81 F1-score using Logistic Regression classifier.
%Temporal features- Temporal characteristics help to understand how individuals publish information on Twitter at different times of the day, such as in the morning or evening.

Most of the related studies, as discussed earlier, were conducted on English datasets. Only one study that of~\citet{9966481} has developed an Arabic suicide dataset translated from an English dataset called ASuiglish. ASuiglish was created by combining several English data sources, which are Twitter datasets provided by ~\cite{10.1093/comjnl/bxab060}, Suicide Notes from the Kaggle dataset~\cite{mashaly_2020}, Victoria Suicide Data~\cite{mashaly_2020} and Suicidal Phrases from Kaggle dataset~\cite{sonu_2020}. The final dataset consists of 1,960 posts with 980 passages per class. The dataset was then pre-processed and cleaned, and abbreviations from the dataset were deleted. After that, the dataset was first translated into Arabic using Google and Microsoft Translate APIs. Then, each entry was examined manually to ensure accuracy and correctness. The dataset was then vectorized using TF-IDF and subsequently fed into the following algorithms: RF, MNB, SVM, and LR. SVM provided a higher accuracy in comparison with the other models. %0.93 accuracy, 0.93 Precision, and 0.92 F1-score. %The resulting dataset was balanced, but due to the heterogeneous sources (Reddit and Twitter), there were lengthier phrases than others. To harmonize segments, manual monitoring was used.

\section{Methodology}
~\label{Sec:Methodology} 
\begin{figure}[h]
   \includegraphics[width=\linewidth]
   {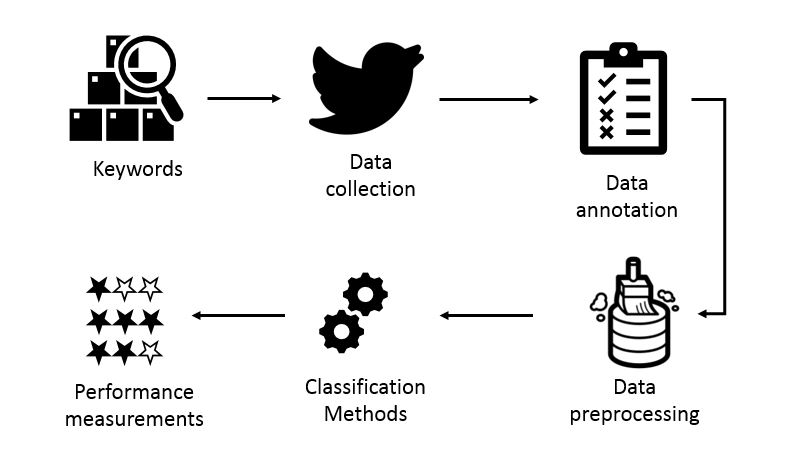}
    \caption{Architecture of Suicide Detection Methodology}
  \label{fig:steps}
\end{figure}

The task of classifying suicide-related posts or blogs aims to identify individuals with suicidal tendencies. Thus, suicidal ideation detection in social media content is often formulated as a supervised learning classification problem. Different machine learning and deep learning techniques have been previously applied to detect suicidality in English and other languages. To identify Arabic suicidal thoughts in social media, our methodology starts with data collection from Twitter using suicidal keywords extracted from previous research. The next step involves annotating and labeling the datasets after removing duplicated tweets, followed by applying feature extraction and preprocessing methods before employing machine learning models. Several machine learning models trained on different sets of feature representation and deep learning models are then employed to identify suicidal content. Last, the performance of the utilized approaches is analyzed.
Figure~\ref{fig:steps} shows the procedure used in this study, which follows most studies discussed in Section~\ref{Sec:related works} 
 and in particular, these studies~\cite{9ji2018supervised,1astoveza2018suicidal,word2ODEA2015183}.
%Our methodology starts with data collection from Twitter using suicidal keywords extracted from previous research. Then, annotating step involves labeling the datasets after removing duplicated tweets. The fourth step is feature extraction which is applied before employing machine and deep learning models. The fifth step is employing machine and deep learning models. The last step is performance analysis.

%This is a primer work further work is in progress.

%The third step, is feature extraction, is applied before employing machine and deep learning models.

\begin{table}[ht]
\centering
%\resizebox{\columnwidth}{!}{%
\begin{tabular}{c c}
\hline
\textbf{Reference} & \textbf{Keywords}\\
\hline
\cite{word1FAHEY2020112960}  & \makecell{I want to kill myself\\ I want to die\\I want to disappear%ابى ابغى اموت اقتل نفسي اختفي
}\\\hline

\cite{word2ODEA2015183} &  \makecell{suicidal; suicide;\\
kill myself;end my life;
\\ never wake up; sleep forever; 
\\want to die; be dead;
\\better off dead; \\ tired of living; \\don't want to be here; \\die alone; 
} \\
\hline

\cite{word3valeriano2020detection}& \makecell{Just want to sleep forever 
\\Kill myself
\\Life is so meaningless
\\Tired of being alone 
\\Don’t want to exist
\\Life is worthless 
\\Don’t want to live 
\\My life is pointless
\\My life is this miserable 
\\My life isn’t worth 
\\Want to be dead 
\\Hate my life 
\\Want to disappear 
\\Hate myself 
\\Suicidal / Suicide Suicida
\\Isn’t worth living
}\\\hline
\cite{9ji2018supervised} &\makecell{I want to end
my life
\\I’m feeling so bad
\\I’m going to kill myself
}\\\hline
\end{tabular}%}
\caption{\label{table1}
Suicidal keywords
}
\end{table}

\subsection{Data collection}
~\label{subSec:Datacollection} 
The first part of this study consists of the data collection process performed using Tweepy. Tweepy is an open-source Python library used to access Twitter API provided by the Twitter developer. An Arabic tweet corpus has been developed from Arabic tweets written in different Arabic dialects (e.g.,~\<عايز اموت , ابى اموت , بدي موت>). Between 23 August 2021 and 21 April 2022, 47,292 tweets were gathered using Arabic suicidal keywords translated from previous English research as shown in Table~\ref{table1}. We have removed duplicated tweets from the dataset. As a result, a total of 5,719 tweets were obtained after removing duplicated tweets.
%The data collection process is performed using Tweepy. Tweepy is an open-source python library used to access Twitter API provided by the Twitter developer. The collected data were gathered based on suicidal Keywords extracted from previous English research shown in table ~\ref{table1}, and then translated to Arabic language. In the interval between 23 August 2021 to 21 April 2022, 47292 tweets were gathered. After deleting duplicated tweets 5720 tweets were left . Different dialect were use e.g.,~\<عايز اموت , ابى اموت , بدي موت> . Therefore, the data were gathered from different Arabic countries.
%20 Feb to 21 Apr

\begin{figure*}[h!]
    \centering
    \begin{subfigure}[t]{0.45\textwidth}
        \centering
        \includegraphics[width=5.5cm,height=5cm]{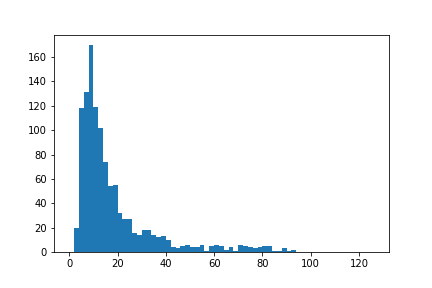}
        \caption{\label{WL1} Average Sentence Lengths for Suicidal Tweets}
    \end{subfigure}%
    \hfill
    \begin{subfigure}[t]{0.45\textwidth}
        \centering
        \includegraphics[width=5.5cm,height=5cm]{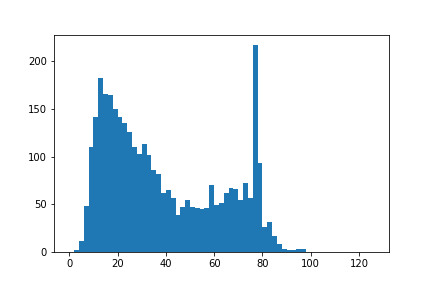}
        \caption{ \label{WL2}Average Sentence Lengths for Non-Suicidal Tweets}
    \end{subfigure}
    \caption{Average Sentence Lengths for  Suicidal and Non-Suicidal Tweets}
\end{figure*}

%\begin{figure}\centering
 %  \includegraphics[width=8cm,height=6cm]
  % {STraninglength.png}
   %\caption{\label{WL1}
   %Average Sentence Lengths for Suicidal Tweets}
 %\end{figure}

%\begin{figure}\centering
%   \includegraphics[width=8cm,height=6cm]
%   {NTraninglength.png}
%   \caption{\label{WL2}
 %  Average Sentence Lengths for Non-Suicidal Tweets}
% \end{figure}

\subsection{Data Annotation}
\label{subSec:data ann} 
All collected tweets were labeled manually by two judges or annotators, one of whom is an expert in psychology, specifically cyberpsychology. Annotators were asked to read the tweets’ text (tweet textual content only) and rate the level of concern of suicide in each tweet. There are two-level of concerns: ‘Suicidal’ tweets labeled with '1', and '0' for ‘Non-Suicidal’ tweets~\cite{word2ODEA2015183}.

%Associate Professor of Psychology at the Graduate School of Education
\begin{itemize}
\item Suicidal: tweet shows serious suicidal ideation; the person expresses a deep and personal intention to commit suicide; e.g., I will kill myself, I want to die, I think of commit suicide\<أقتل نفسي , نفسي اموت , افكر انتحر>. In contrast, a  suicide risk will not be considered if the content relevent to some type of conditional event e.g., I will kill myself if my team doesn't win today, I wish to die once the doctor says quiz, I'm thinking of committing suicide if the third part didn't come out, \<حقتل نفسي لو ما فاز الاتحاد اليوم \\، اتمنى اموت اول ما يقول الدكتور كويز \\، افكر انتحر لو مانزلوا جزء ثالث> unless this event was a serious suicidal risk factor e.g., abuse, drug use, or bullying e.g., Her words are beyond my power, I want to die, \<كلامها تجاوز قوتي ابى اموت>.
%; the occurrence of suicide appears to be imminent e.g.,  versus
%\item Possibly concerning: the default category.
\item Non-Suicidal: no fair evidence to indicate that there is a possibility of suicide; 
e.g. I lost my mind when I singed myself a leader, I thing of committing suicide , \<وين كان عقلي يوم سجلت  ليدر؟ 
افكر انتحر>
\end{itemize} %cite {Suicidal Behavior Detection on Twitter Using Neural Network}
%وين كان عقلي يوم سجلت رقمي ليدر بكل المواد؟افكر انتحر
%2 undeterment: ابي طريقه اموت فيها وادخل الجنه دايركت
%عندي تقارير وعرضين لازم اسلمهم رمضان تحسون اسويهم اقوم اذبح نفسي وافتك

\begin{figure*}[t!]
    \centering
    \begin{subfigure}[t]{0.45\textwidth}
       % \centering
        \includegraphics[width=5.5cm,height=5cm]{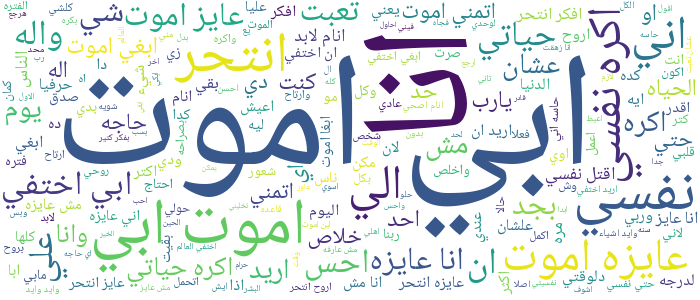}
        \caption{\label{figW1} Word cloud for Suicidal tweets}
    \end{subfigure}%
    \begin{subfigure}[t]{0.45\textwidth} \hfill
     %   \centering
        \includegraphics[width=5.5cm,height=5cm]{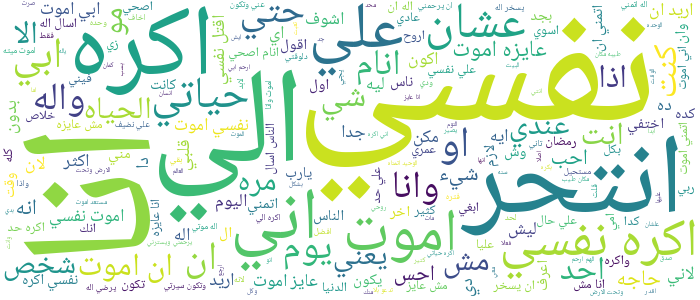}
        \caption{\label{figW2} Word cloud for Non-Suicidal tweets}
    \end{subfigure}
    \caption{Word Cloud for Suicidal and Non-Suicidal Tweets}
\end{figure*}

%\textwidth
%\begin{figure}\centering
%   \includegraphics[width=7.5cm,height=5cm]
%   {wordcloudhSD.png}
%   \caption{\label{figW1}
%   Word cloud for Suicidal tweets}
%\end{figure}

%\begin{figure}\centering
%   \includegraphics[width=7.5cm,height=5cm]
%   {wordcloudhND.png}
%   \caption{\label{figW2}
%   Word cloud for non-Suicidal tweets}
%\end{figure}

%\begin{figure}\centering
%   \includegraphics[width=7.5cm,height=5cm]
%   {wordcloudh1not0predrop.png}
%   \caption{\label{figW2}
%   Word cloud for Dataset}
%\end{figure}
\begin{figure*}[t!]
    \centering
    \begin{subfigure}[t]{0.45\textwidth}
       % \centering
        \includegraphics[width=5.5cm,height=5cm]{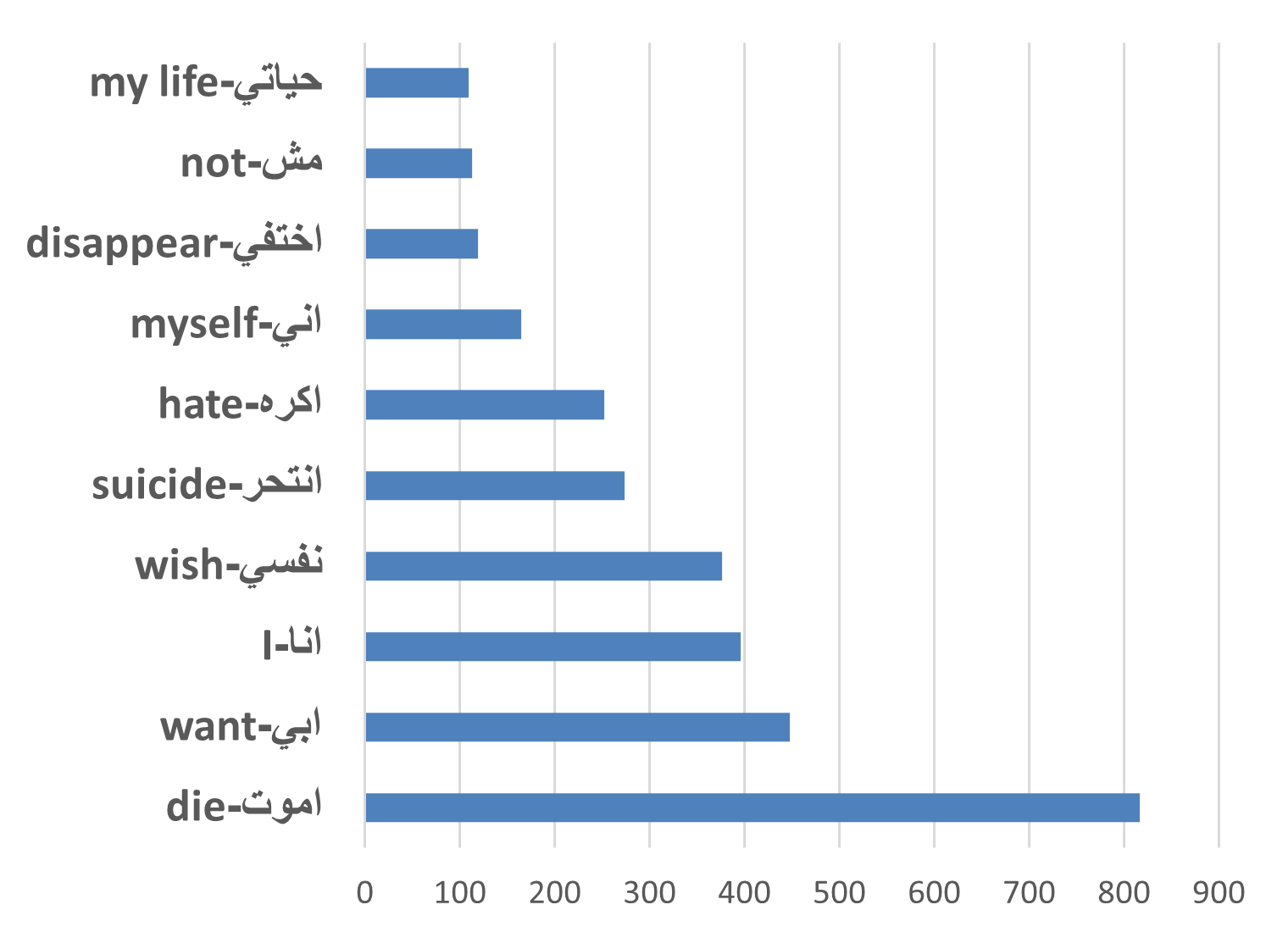}
        \caption{\label{Common1} Common Words Found in Suicidal Tweets (Without Stop Words)}
    \end{subfigure}%
    \begin{subfigure}[t]{0.45\textwidth} \hfill
     %   \centering
        \includegraphics[width=5.5cm,height=5cm]{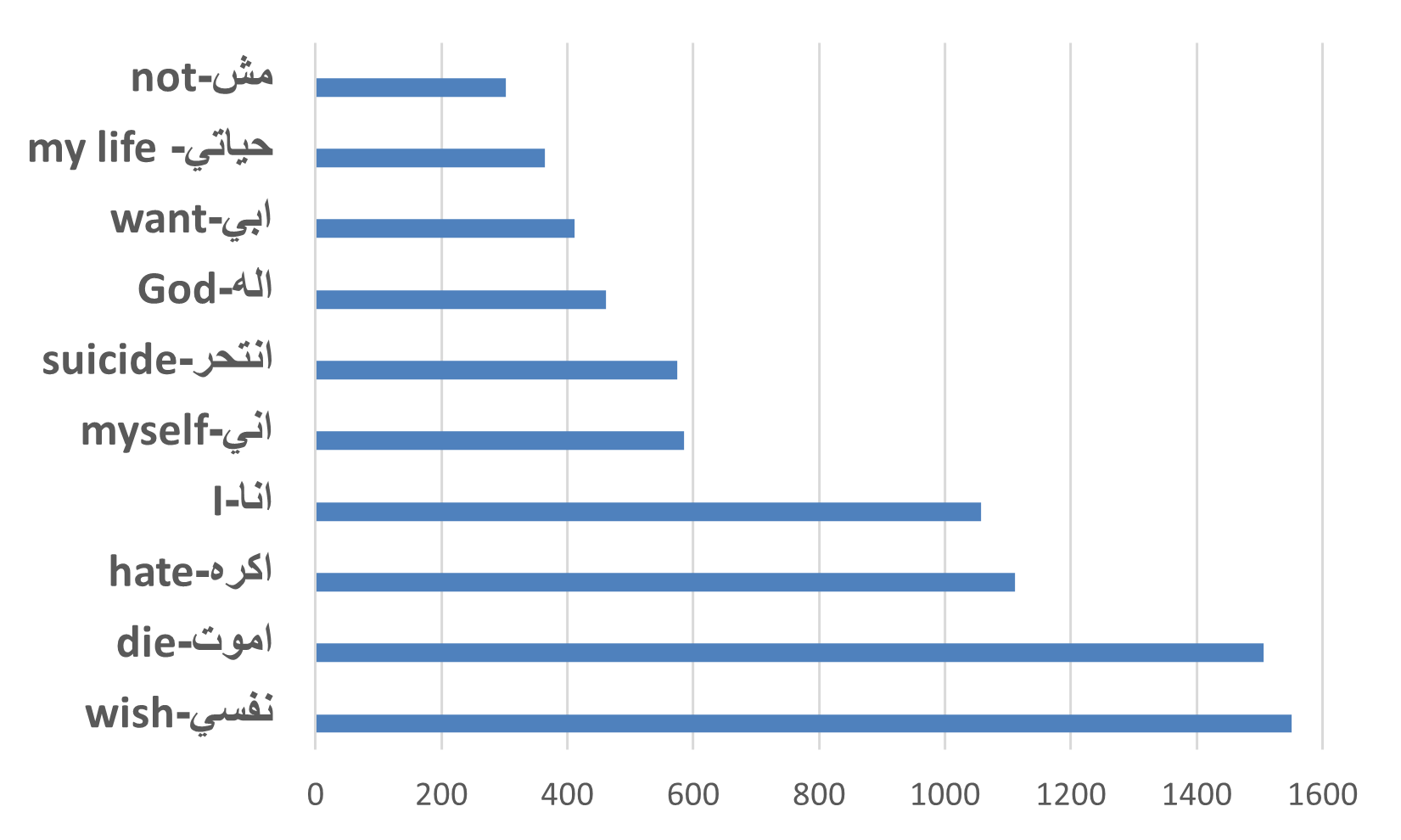}
        \caption{\label{Common2} Common Words Found in non-Suicidal Tweets (Without Stop Words)}
    \end{subfigure}
    \caption{ 
   Common Words Found in Suicidal and Non-Suicidal Tweets (Without Stop Words)}
\end{figure*}

%\begin{figure*}\centering
%   \includegraphics[width=13cm,height=10cm]
%   {S10mot.png
%   %plot4.png   
%   }
%   \caption{\label{Common}
%   Common Words Found in Suicidal Tweets (Without Stop Words)
%   }
%\end{figure*}

%\begin{figure*}\centering
 %  \includegraphics[width=13cm,height=10cm]
 %  {Not10mot.png
   %plot4.png   
 %  }
 %  \caption{\label{Common}
 %  Common Words Found in Non-Suicidal Tweets (Without Stop Words)
 %  }
%\end{figure*}

\begin{figure*}[t!]
    \centering
    \begin{subfigure}[t]{0.45\textwidth}
       % \centering
        \includegraphics[width=5.5cm,height=5cm]{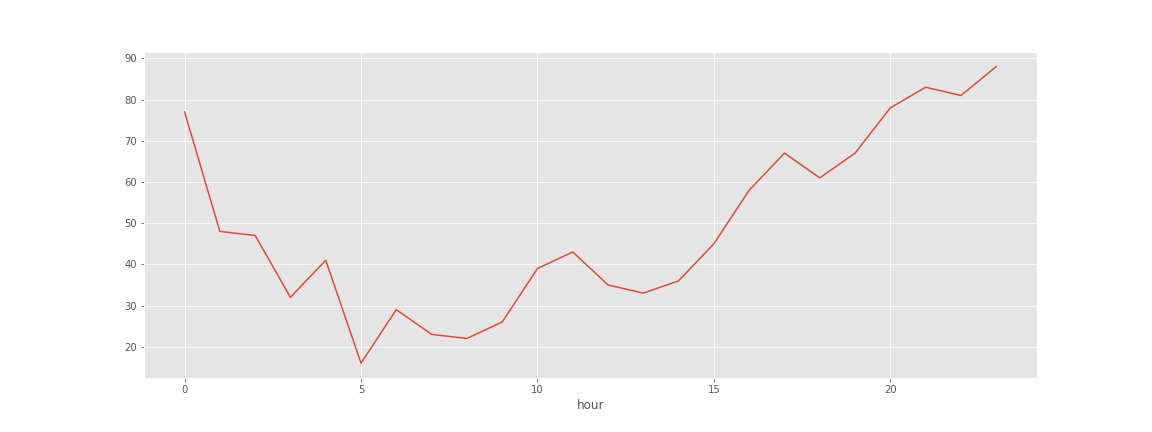}
        \caption{\label{hours1} Trends of active hours for Suicidal tweets.}
    \end{subfigure}%
    \begin{subfigure}[t]{0.45\textwidth} \hfill
     %   \centering
        \includegraphics[width=5.5cm,height=5cm]{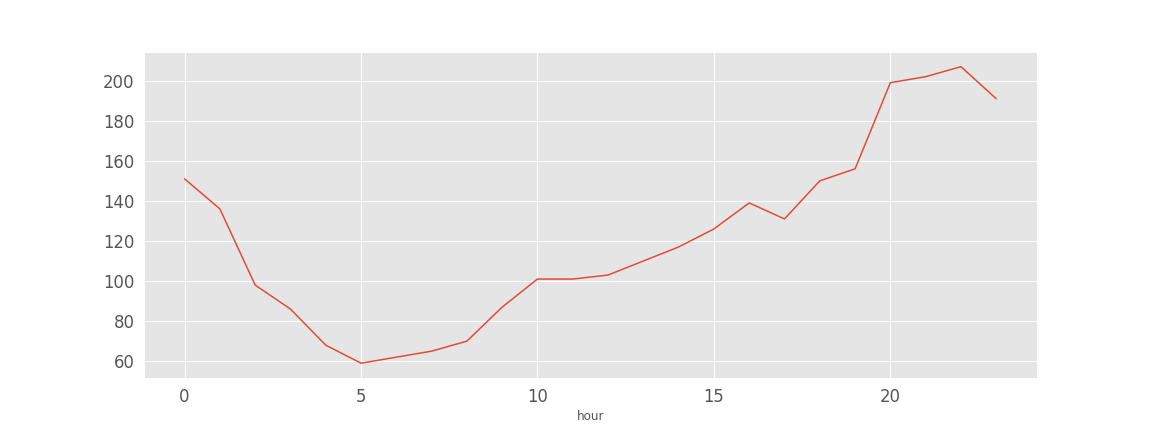}
        \caption{\label{hours2} Word cloud for Non-Suicidal tweets}
    \end{subfigure}
    \caption{ Trends of active hours for Non-Suicidal tweets}
\end{figure*}

After annotating each tweet, 1,426 tweets were labeled as Suicidal tweets and 4,293 as non-Suicidal tweets. Table~\ref{table2} shows dataset classes and their weights. The first class is suicide ideation tweets (Suicidal), and the second class indicates that there is no potential suicidal intention (Non-Suicidal). It has been noticed that most suicidal tweets are short, ranging from 2 to 20 words, as shown in Figure~\ref{WL1}, while non-Suicidal tweets vary from 2 to 80 words, as shown in Figure~\ref{WL2}. The most frequent words used in each class are shown in word-cloud representations in Figure~\ref{figW1} and~\ref{figW2}, and as shown in Figure~\ref{Common1} and~\ref{Common2}, the most frequent word used in a Suicidal tweet is \<"اموت">,  it's been used and repeated more than 800 times. As shown in Figure~\ref{hours1} and ~\ref{hours2}, the peaks of the suicidal tweets show up after ten o’clock at night and continue to increase, and then there is a clear decrease at five o’clock in the afternoon.
%As its shown in ~\ref{WL1} most of suicidal tweets are short 

\begin{table}[ht]
\centering
\begin{tabular}{ll}
\hline
\textbf{Class} & \textbf{\%of Datasets}\\
\hline
Suicidal  & \makecell{25\% }\\
%Non- suicidal & No evidence of possible suicidal intent & 75\% \\
Non-Suicidal &  \makecell{75\%} \\
\hline
\end{tabular}
\caption{\label{table2}
Labelled tweets and their weights
}
\end{table}

\subsection{Annotators Agreements}
~\label{subSec:kappa}

After labeling the dataset, the inter-rater agreement was assessed using Cohen’s Kappa to measure the pairing agreement between a set of annotators who make classification judgments after accounting for the likelihood of chance agreement. It’s calculated using the following formula:
\begin{equation}
K = \frac{P(A) -  P(E)}{1 - P(E) }%P(A) - P(E) /1 - P(E) 
\end{equation}

Where P(A) is the number of times the annotators agree, P(E) is the number of times expected from them to agree by chance and is calculated in accordance with the above-mentioned intuitive argument~\cite{kappasidney1957nonparametric}. The value of $K$ equals zero if there is no agreement beyond the level expected by chance. In the case of complete agreement, $K$ equals one. In this research, the inter-rater agreement between the two annotators of the dataset equals 0.978, which indicates almost perfect agreement and reliability. Table~\ref{kappaT} shows the quantified agreement between the two annotators.    %the labelling process was performed by two annotators and based on the calculation made on Table~\ref{kappaT}, the value of K equals 0.978, which eliminates bias and indicates perfect agreement.  

\begin{table}[htp!]
\centering

\begin{tabular}{|c|c|c|c|}
\hline
       & \textbf{A}  & \textbf{B} &\textbf{Total}         \\ \hline
\textbf{A}  & 4258 & 8 & 4293 \\ \hline
\textbf{B}  & 39 & 1387 & 1426 \\ \hline
\textbf{Total} & 4324 & 1395 & 5719\\ \hline
\end{tabular}
\caption{\label{kappaT} Quantified agreement with kappa results}
\end{table}

%\begin{figure*}\centering
%   \includegraphics[width=\textwidth,height=8cm]
%   {maxhourSall-.png
%   %plot4.png   
%   }
%   \caption{\label{hours}
%   Trends of active hours for Suicidal tweets.
%   %Frequency of tweets during day hours
%   }
%\end{figure*}

%\begin{figure*}\centering
%   \includegraphics[width=\textwidth,height=8cm]
%   {maxhourUnSall.png
%   %plot4.png   
%   }
%   \caption{\label{hours}
%   Trends of active hours for Non-Suicidal tweets.
%   %Frequency of tweets during day hours
%   }
%\end{figure*}

%\subsection{Data Pre-processing}

\subsection{Feature Extraction}
\label{subSec:feature} 
Feature extraction is a critical task in machine learning models and natural language processing problems that require transforming text to representative numerical representation. %We used the text feature as an efficient feature set to distinguish Suicidal tweets from Non-Suicidal tweets.
Several features were used in this study, including word frequency features (i.e., BOW and TF-IDF), (generic) word embedding features (i.e., word2vec, FastText), and contextualized word embedding used in the deep learning models (i.e., AraBert, AraELECTRA, and AraGPT2).
\begin{itemize}

%CountVectorizer
 \item Bag of Words (BOW): is a standard word frequency model that converts text documents into numerical representations based on their occurrence in the document. The BOW model represents each document in the dataset as a feature vector of |V| where V is the corpus's unique words or vocabulary, and each unit in the vector represents the presence of known words. 
 Bag of Words is one of the most commonly used text modeling approches in natural language processing and information retrieval. The BOW  model also has several limitations, such as ignoring grammatical structure and word orders as well as representing textual data in a very sparse representation~\cite{deepa2019sentiment}.
 %\item Bag of Word (BOW): is a standard word frequency model that converts text document into numbers (each token in the data is converted into a feature). Each word or token in the dataset is set to a value, and each value is set to the number of times the word appears. %The classifier can then receive this multi-set of words as input. %This process that represent a dataset as a bag of words vector is known as tokenization. 
 %Bag of Word is one of the most often used methodologies in text modeling in natural language and information retrieval tasks. The BOW approach also has some drawbacks, such as neglecting grammatical structure and word ordering and representing textual data in a relatively sparse representation~\cite{deepa2019sentiment}.
%As a whole it transforms into array of features and create matrix. It count the repetition of each word in the document

\item TF-IDF: is a word frequency model that weights word counts by the number of times a word appears in the corpus. TF-IDF is used to improve the performance of the standard word frequency approach by giving less weight to common words that frequently appear in the documents using the formula described below.

\begin{equation}
  TF\textrm{--}IDF = TF_{i,j}*IDF_i  
\end{equation}
where $TF_{i,j}$ is the number of times the word ($i$) occurs in the document($j$) divided by the total number of words in the document ($j$) and $IDF_i$ is the log of the total number of documents in the corpus divided by the number of documents containing the word ($i$). 
%\begin{itemize}
%\item TF = number of times the term occurs in
%the text divided by the total number of words in the text,
%\item IDF = log of the total number of documents divided by the number of documents with the word in it.
%\item TF-IDF can be calculated by,\newline
% \cite{deepa2019sentiment}
%\end{itemize}

All n-grams (unigram, n-gram, and character level n-gram) features are normalized by $TF\textrm{--}IDF$ value to determine their relative importance in the corpus.

\begin{itemize}

\item Unigram: The basic type of textual feature, which consists of an individual word $TF\textrm{--}IDF$ value.
\item N-grams: A combination of word bigram and trigram.
\item Character n-gram: %contiguous sequences of characters extracted from a text corpus. 
%Designed to extract character-level n-gram features (sequences of 2 or 3 characters) from the input text.
%The Arabic alphabet consist of 28 letters in addition to the space character.
based on single characters, or sequences of characters, rather than whole words or sentences. In order to calculate the frequency of each character in each document, each character was treated as a "word." As a result, a matrix is created where each row corresponds to a document and each column corresponds to a character. The standard TF-IDF formula is then applied to this matrix to obtain a weighted representation of the character level features. This technique is particularly important for morphologically rich languages such as Arabic, as it can identify the morphological components of words. It is also useful for identifying word misspellings and alternative spellings that are common in online communications~\cite{alsafari2020hate}. The character n-gram has achieved state-of-the-art performance on several text classification tasks~\cite{zhang2015character}

\end{itemize}
\item Word embedding features:
Word embedding has achieved a significant advancement in numerous areas of natural language processing (NLP) in recent years, such as sentiment analysis, topic segmentation, text mining and recommendation~\cite{xu2018incorporating,Li_Shah_Liu_Nourbakhsh_2017, naili2017comparative}. In this research, two common word embedding models (Word2vec and Fasttext) were used. A study by~\citet{kaibi2019comparative} showed that Fasttext, followed by Word2vec, gave the best techniques for Twitter Sentiment Analysis~\cite{kaibi2019comparative}. 

\begin{itemize}
    \item Word2vec: is one of the most popular word embedding models created by ~\citet{mikolov2013efficient}. It is based on two neural network model architectures (Skip-gram and Continuous bag-of-words (CBOW)). Skip-gram uses the word as input and predicts the surrounding words as output, while CBOW uses the context words to predict the word. The Arabic Word2vec (AraVec) model provides powerful and free-to-use word embeddings for Arabic natural language processing research. It is built on three different Arabic content domains including Tweets, World Wide Web pages, and Wikipedia Arabic articles, and contains over $3.3$ billion tokens~\cite{SOLIMAN2017256}. The Arabic Word2vec model has proven to be effective in capturing semantic and semantic relationships between Arabic words and has achieved the state-of-the-art results in various tasks of processing Arabic natural languages~\cite{doi:10.1137/1.9781611974010.66,HEIKAL2018114}.
%such as AraVec and AWE,
 \item FastText: is an extension of the Skip-gram model. %~\cite{grave2018learning}
Each word in FastText is represented by a sum of n-gram vectors~\cite{athiwaratkun2018probabilistic}. It was adapted for the Arabic language to create word embeddings for sentiment analysis tasks in Arabic. The Arabic FastText model outperformed other Arabic word embedding models on various Arabic NLP tasks, such as sentiment analysis and named entity recognition~\cite{computers12060126}.
\end{itemize}

\end{itemize}
%Figure ~\ref{figW1} illustrates the top most frequent word used in Suicidal tweet, while figure ~\ref{figW2} illustrates the top most frequent word used in non-Suicidal tweet

\subsection{Models and Hyperparameters}%or \subsection{Classification}
~\label{subSec:training} 
%In this experiment, the performance of seveal/; machine/; and/; deep/; leaning/; algothim/; wee/; use.

%In this experiment, the performance of deep learning models AraBert, AraELECTRA, and AraGPT2 were compared with five popular machine learning models trained on different sets of textual features representation to classify Arabic suicidal tweets. The five machine learning models used in this research are Naïve Bayes (NB), Support vector machine (SVM with gamma=1 and c =10), K-Nearest Neighbor (with k=30 and Euclidean distance as distance measure), Random forest(RF), and Extreme Gradient Boosting (XGBoost) with extreme gradient (boosting=6) to classify Arabic suicidal tweets. %AraBert is a pre-trained Bidirectional Encoder Representation from Transformer (BERT) model developed by Google for the Arabic language.\cite{berthttps://doi.org/10.48550/arxiv.1810.04805}

    In this experiment, the performance of deep learning models AraBert, AraELECTRA, and AraGPT2 was compared with five popular machine learning models trained on different sets of textual feature representations to classify Arabic suicidal tweets. The five machine learning models used in this research were Gaussian NB, SVM, KNN, RF, and XGBoost. Gaussian NB classifiers were used with smoothing hyperparameters. Smoothing is a technique used to address the issue of zero probability (no likelihood) in the GNB. The smoothing parameter in our study takes values between 1E-11 and 1E-7 according to~\cite{john2013estimating,informatics8040079}. %The three most important SVM hyperparameters are kernel, C, and gamma. The kernel has the following values: "linear," "poly," "rbf," "sigmoid," and "precomputed." By default, scikit-learn uses "rbf." We have the option of using a "linear" or "poly" kernel if the dataset is linearly separable. In the case where the data cannot be linearly separated, we can choose the "rbf" kernel as in our case. The two most important hyperparameters that can be used to train the best SVM model using an RBF kernel are gamma and C values. Here, in our study, we used

The SVM model was implemented with RBF kernel, with C values ranging from 1 to 10 and gamma from 0.1 to 1 as these hyperparameters gave the best performance on the evaluation set. This range was chosen based on previous research on Arabic Sentiment Analysis for comprehensive comparative analysis of various hyperparameter tuning techniques~\cite{informatics8040079}. 
For the KNN algorithm, the Euclidean distance was used for similarity measure andThe k neighbors’ values were selected in the range between 1 and 31; BOW’s best k value was n\_neighbors=2, and for unigram, it was n\_neighbors=3, while with n-gram and character n-gram, it was 30. For the Random Forest (RF) model, the parameters "max\_features" were set to "log2" and "n\_estimators" to 1000 as they result in greater accuracy compared to the default values, as also reported in~\cite{informatics8040079}. % The number of trees in the model is specified by the n\_estimators parameter. This parameter's default value is 10 means that the random forest will create 10 separate decision trees. The max\_features, as the name indicts, is the maximum feature to be taken into account while splitting a node, and the default value 'auto' is used. 
In our case, the "max\_features" was chosen based on the highest performance from ['auto,' 'sqrt,' 'log2'], and "n\_estimators" was chosen from [100,200,300,1000]. 
%XGBoost is a scalable and effective use of gradient-boosting technology. 
As RF, XGBoost has "n\_estimators," and its default value is set to 100 and as its used by~\cite{AMOUDI202212511} %(Amoudi, Ghada et al.,2022) 
for Arabic text classification, the number of estimators was set to 200 and 300. %~\cite{AMOUDI202212511}.

%AraBERT was developed based on the BERT model~\cite{antoun2020arabert,el2022arabert}.

%the batch size used during fine-tuning of the AraBERT model is set to 32.

%AraBERT is a pre-trained Arabic language model developed by the Arabic Language Technologies group at the King Abdullah University of Science and Technology (KAUST). It is based on the BERT architecture %and consists of 12 transformer layers, 768 hidden units, and 12 self-attention heads. 

AraBERT is a pre-trained Arabic language model, trained utilizing a masked language modeling objective on a large corpus of Arabic text data. It has been shown to deliver state-of-the-art results on a variety of Arabic language processing tasks~\cite{antoun2020arabert}. Other pre-trained Arabic language models that have been developed recently in addition to AraBERT are AraELECTRA and ARAGPT2. AraELECTRA is a variant of the ELECTRA model, which pre-trains a language model using a generator-discriminator architecture. It has been demonstrated to achieve competitive performance on a number of Arabic natural language processing tasks~\cite{antoun-etal-2021-araelectra}. ARAGPT2 is a variant of the GPT-2 model, which is a transformer-based language model. ARAGPT2 was pre-trained on a large corpus of Arabic text data and has demonstrated state-of-the-art performance on several Arabic language processing tasks, including text generation and language modeling~\cite{antoun-etal-2021-aragpt2}. A batch size of 16, epoch size of 5, the value of adam\_epsilon is set to 1e-8, and a learning rate of 2e-5 were used for fine-tuning the models.

    %per_device_train_batch_size = 16, # up to 64 on 16GB with max len of 128
    %per_device_eval_batch_size = 128,
    %gradient_accumulation_steps = 2, adam_epsilon = 1e-8,

% BERT uses a masked language model (MLM) objective, which allows the left and right fuse context. Unlike other language representation models that uses only left to right language model pretraining [29]. BERT model is utilized in wide range of NLP tasks, such as question answering, sentiment classification and provide the state of art accuracy [29, 30].  BERT enhanced the ability of word embedding model generalization. It can fully express the character level, word level, sentence level, and sentence relationship [6]. state-of-the-art models for a wide range of tasks, such as question answering and language inference

\subsection{Performance Evaluation}
In order to evaluate the performance of our system and assess the capacity of the models used in this experiment, we considered the measurements of precision, recall, F1 score (F1), and accuracy defined as follows:

%Using the following formula:
%\begin{itemize}
\begin{equation}
     Precision =\frac{TP}{TP+FP}% TP/ (TP + FP)
     \end{equation}

\begin{equation}
     Recall = \frac{TP}{TP+FN}%TP/(TP + FN)   
    \end{equation}
\begin{equation}
 F1 = \frac{2*(Precision * Recall)}{Precision + Recall}%2*(precision * recall)/(precision + recall)
\end{equation}
\begin{equation}
     Accuracy = \frac{TP+TN}{TP+TN+FP+FN}%TP+TN/(TP+TN+FP+FN)
 \end{equation}
%\end{itemize}
Where $TP$ denotes the true positives, or model correct predictions of the positive class; $FP$ denotes the false positives, or model incorrect predictions of the positive class; $TN$ denotes the true negatives, or model correct predictions of the negative class; and $FN$ denotes the false negatives, or model incorrect predictions of the negative class.

%\begin{itemize}
 %   \item TP = true positives,the model correctly predicts the positive class.
 %   \item FP = false positives,the model incorrectly predicts the positive class.
  %  \item TN = true negatives,the model correctly predicts the negative class.
  %  \item FN = false negatives,the model incorrectly predicts the negative class.
%\end{itemize}
%cite Analyzing Tweets For Predicting Mental Health States Using Data Mining And Machine Learning Algorithms

%plot2Classes+Copy of gridsearchCV.ipynb

\section{Result analysis}
\label{Sec:result} 
\begin{figure*}\centering
   \includegraphics[width=\textwidth,height=15cm]
   {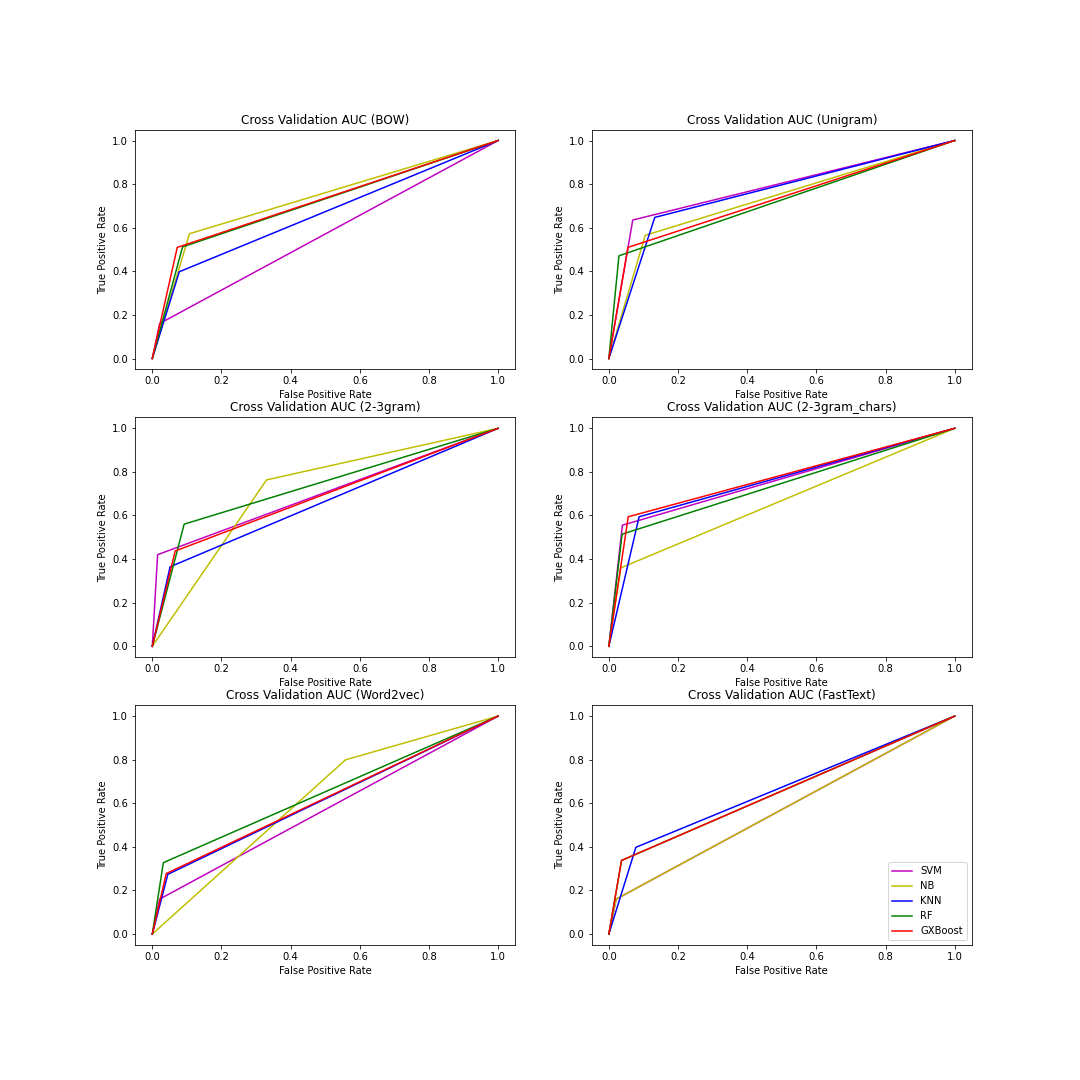
   %plot4.png   
   }
   \caption{\label{ROC}
   ROC Curves depicts the performance  obtained for Machine Learning algorithm}
\end{figure*}

\begin{figure*}\centering
   \includegraphics[width=\textwidth,height=4cm]
   {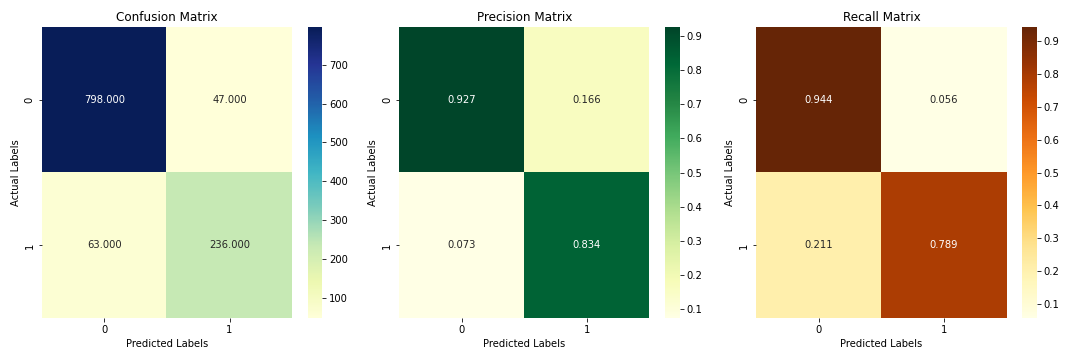}
   \caption{\label{conarabert}
   Confusion Matrix for Best performing model (AraBert)}
\end{figure*}

The BOW, TF-IDF with different levels (unigram, bigram and trigram, character n-gram), and word embedding features were used for representation. %As it's shown in Table 3, TF-IDF CharLevel gives better performance
Five machine learning models were examined and compared against deep learning models, which are Naïve Bayes, Support vector machine, K-Nearest Neighbors, Random forest, and XGBoost. Accuracy, Precision, Recall, and F1-Score were used for evaluation. Table~\ref{table3} shows the comparison of classifiers. The results show that character n-gram TF-IDF features gave the best performance (accuracy, precision, recall, and F1-score ) in all the machine learning models in comparison with the BOW model and word level unigram, bigram, and trigram. The highest accuracy in the machine learning model was 86\% obtained using SVM and RF using the character n-gram feature. Figure~\ref{ROC} presents the ROC curve (receiver operating characteristic curve) obtained for the five algorithms (Naïve Bayes, SVM, k-nearest neighbors, Random Forest, and Xgboost). 
%Figure \ref{ROC} depicts the performance  obtained for the five algorithms (Naïve Bayes, SVM, Random Forest and Xgboost).
%
The results also show that the performance of the deep learning models was better than all the machine learning models in identifying suicidal ideation in Arabic tweets. The AraBert model gave the best performance in comparison with all other machine learning and deep learning models with 88\% precision, 89\% recall, 88\% F1-score, and 91\% accuracy in classifying suicidal thoughts in Arabic tweets. The confusion matrix of AraBert is shown in Figure~\ref{conarabert}.

\begin{table}[ht]
%\label{bigtable}
\centering
\begin{tabular}{llllll}
\hline
\textbf{Classifier} & \textbf{Feature} & \textbf{Precision} & \textbf{Recall}& \textbf{F1-score} & \textbf{Accuracy} \\
\hline
NB & \makecell{Bag of word\\ Unigram\\ N-Gram\\CharLevel* \\ Word2Vec \\ Fasttext} & 
\makecell{73\% \\75\%\\ 73\%\\ \textbf{81\%} \\ 77\% \\76\%}& 
\makecell{73\% \\ 72\%\\ \textbf{75\%}\\ 72\%\\ 54\% \\ 62\%}&
\makecell{73\% \\ 74\%\\  74\%\\\textbf{75\%}\\ 51\% \\ 63\%}&
\makecell{81\% \\ 82\%\\   80\%\\\textbf{84\%}\\ 76\% \\ 78\%} \\
% without gridsearch
%accuracy   0.59      
%macro avg       0.65      0.62      0.58   
%Fasttext without gridsearch: accuracy= 0.71 macro avg       0.72      0.67      0.68      1144
%GaussianNB(var_smoothing=1.0)

\hline
%0.73      0.73      0.73  0.81  
%MultinomialNB(alpha=1e-07)
% 0.75      0.72      0.74  0.82  MultinomialNB(alpha=1e-07)
%0.73      0.75      0.74   0.80 MultinomialNB(alpha=1e-06)
%0.81      0.72      0.75   0.84    MultinomialNB(alpha=0.001)

SVM & \makecell{Bag of word\\  Unigram\\ N-Gram\\CharLevel* \\ word2vec \\Fasttext} & 
\makecell{82\% \\84\%\\ \textbf{86\%}\\ 85\%\\ 78\% \\77\%}&
\makecell{\textbf{78\%} \\ 76\%\\ 72\%\\ 76\%\\ 66\% \\72\%}&
\makecell{\textbf{79\%} \\ \textbf{79\%}\\76\%\\\textbf{ 79\%}\\ 68\% \\74\%}&
\makecell{\textbf{86\%} \\ 86\%\\85\%\\ \textbf{86\%}\\  80\% \\81\%}\\
%word2vec {'C': 10, 'class_weight': 'balanced', 'gamma': 1, 'kernel': 'rbf', 'probability': True}SVC(C=10, class_weight='balanced', gamma=1, probability=True)
%fastext without: accuracy=0.81  macro avg       0.69      0.77      0.71 
%SVC(C=10, class_weight='balanced', gamma=0.1, probability=True)
\hline
%C and gamma
%0.82      0.78      0.79  0.86 Grid
%SVC(C=10, class_weight='balanced', gamma=0.1, probability=True)
%0.84      0.76      0.79  0.86  Grid
%SVC(C=10, class_weight='balanced', gamma=1, probability=True)
%0.88      0.57      0.57  0.79
%0.86      0.72      0.76  0.85 grid
%SVC(C=1, class_weight='balanced', gamma=1, kernel='linear', probability=True)
%0.85      0.76      0.79  0.86
%{'C': 10, 'class_weight': 'balanced', 'gamma': 1, 'kernel': 'rbf', 'probability': True} SVC(C=10, class_weight='balanced', gamma=1, probability=True)SVC(C=10, class_weight='balanced', gamma=1, probability=True)
KNN & \makecell{Bag of word\\  Unigram\\ N-Gram\\CharLevel*\\ Word2Vec \\ Fasttext} & 
\makecell{63\% \\75\%\\ 75\%\\ \textbf{78\%}\\ 64\% \\71\%}&
\makecell{68\% \\ 76\%\\ 69\%\\ 75\% \\ 71\%\\ \textbf{77\%}}&
\makecell{60\% \\ 75\%\\  71\%\\ \textbf{76\%} \\ 65\% \\73\%}&
\makecell{63\% \\ 81\%\\81\%\\ \textbf{84\%} \\77\% \\81\%}\\
%word2vec reported result wihout grid
%FastText accuracy=0.80, macro avg       0.77      0.67      0.69    
\hline
%0.63      0.      0.60    0.63   KNeighborsClassifier(n_neighbors=2)
% 0.75      0.76      0.75   0.81  KNeighborsClassifier(n_neighbors=30)
%0.75      0.69      0.71   0.81   KNeighborsClassifier(n_neighbors=3)
%0.78      0.75      0.76   0.84  KNeighborsClassifier(n_neighbors=30)
RF & \makecell{Bag of word\\  Unigram\\ N-Gram\\CharLevel*\\ Word2Vec \\Fasttext} & 
\makecell{77\%\\ 83\%\\ 77\%\\ \textbf{85\% }\\ 81\% \\69\%}&
\makecell{76\% \\ 72\%\\ 73\%\\74\% \\ 65\% \\\textbf{81\%}}&
\makecell{\textbf{77\%} \\ 75\%\\ 75\%\\ \textbf{77\%} \\ 67\% \\ 72\%}&
\makecell{83\% \\ 85\%\\ 85\%\\\textbf{86\%} \\ 81\% \\82\%}
%RandomForestClassifier(criterion='entropy', max_features=2, min_samples_split=6)
%FAsttext repotered result without gridsearch RandomForestClassifier(criterion='entropy', max_features=3) accuracy= 0.81  macro avg       0.81      0.67      0.70      1144
\\\hline
%0.77      0.76      0.77    0.83
%0.83      0.72      0.75    0.85
%0.77      0.73      0.75    0.83
%0.85      0.74      0.77   0.86
XGBoost & \makecell{Bag of word\\  Unigram\\ N-Gram\\CharLevel*\\ Word2Vec \\ Fasttext} & 
\makecell{79\% \\\textbf{83\%}\\ 79\%\\ 82\% \\ 64\%\\ 81\%}&
\makecell{67\% \\ 69\%\\ 60\%\\\textbf{75\%} \\ 78\% \\ 68\%}&
\makecell{69\% \\ 72\%\\62\%\\78\% \\ 66\% \\ \textbf{80\%}}&
\makecell{82\% \\ 84\%\\80\%\\\textbf{85\%}\\ 79\% \\70 \%}\\
\hline
%fasttext with gridsearch XGBClassifier(colsample_bytree=0.7, learning_rate=0.5, max_depth=6, min_child_weight=11, missing=-999, n_estimators=5, seed=1337,  silent=1, subsample=0.8) accuracy        =0.80 macro avg       0.75      0.68      0.70      1144
%0.79      0.67      0.69   0.82
%0.83      0.69      0.72   0.84
%0.79      0.60      0.62   0.80
%0.82      0.75      0.78   0.85
%CNN & \makecell{glove} %\makecell{Count Vectors\\ TF-IDF\\ N-Gram} &    \makecell{50\%}& \makecell{38\%}& \makecell{43\%}&\makecell{76\%} \\\hline
%RNN & \makecell{glove} %\makecell{Count Vectors\\ TF-IDF\\ N-Gram} &    \makecell{50\%}& \makecell{38\%}& \makecell{43\%}&\makecell{76\%} \\\hline
%LSTM & \makecell{glove} %\makecell{Count Vectors\\ TF-IDF\\ N-Gram} &    \makecell{50\%}& \makecell{38\%}& \makecell{43\%}&\makecell{76\%} \\\hline

%LSTM & \makecell{Word2Vec} %\makecell{Count Vectors\\ TF-IDF\\ N-Gram} 
%&    \makecell{64\%}& \makecell{61\%}& \makecell{62\%}&\makecell{75\%} \\
%\hline
%new
%aubmindlab/bert-base-arabertv02-twitter: 0.88      0.89      0.88  0.91
%UBC-NLP/MARBERT:                     0.91      0.87      0.88  0.92
% aubmindlab/araelectra-base-generator: 0.85      0.84      0.85  0.88
%aubmindlab/AraGPT22-mega-detector-long: 0.87      0.85      0.86  0.89

%old
%aubmindlab/bert-base-arabertv02-twitter: 0.88      0.86      0.87  0.90
% aubmindlab/araelectra-base-generator: 0.86      0.82      0.84  0.88
%aubmindlab/AraGPT22-mega-detector-long: 0.85      0.81      0.83  0.87
AraBert & \makecell{-} %\makecell{Count Vectors\\ TF-IDF\\ N-Gram} 
&    \makecell{\textbf{88\%}}& \makecell{\textbf{89\%}}& \makecell{\textbf{88\%}}&\makecell{\textbf{91\%}} \\
AraELECTRA& \makecell{-} %\makecell{Count Vectors\\ TF-IDF\\ N-Gram} 
&    \makecell{85\%}& \makecell{84\%}& \makecell{85\%}&\makecell{88\%} \\
AraGPT2& \makecell{-} %\makecell{Count Vectors\\ TF-IDF\\ N-Gram} 
&    \makecell{87\%}& \makecell{85\%}& \makecell{86\%}&\makecell{89\%} \\
\hline
\end{tabular}
\caption{\label{table3}
 Performance analysis of different classifiers in classifying Suicidal Tweets.* indicates best performing feature extraction technique in all the machine learning models.
}
\end{table}

\section{Conclusion}
~\label{Sec:conclusion} 
 
This study demonstrates the potential of machine learning algorithms and deep learning models, including pre-trained transformer deep learning models (AraBert, AraELECTRA, and AraGPT2), in detecting suicidal ideation in Arabic tweets. To investigate the viability of our proposed approach, a novel Arabic suicidal dataset was developed and annotated in this study. Our experiments show that pre-trained transformer deep learning models outperformed traditional machine learning models trained on different sets of features. The best performance was obtained using AraBert, achieving an 88\% precision, 89\% recall, 88\% F1-score, and 91\% accuracy in identifying Arabic tweets that express suicidal ideation. This study provides a foundation for future research in using machine learning to detect and prevent self-harm and suicide at an early stage. Future work will focus on increasing the size of the dataset and exploring other machine and deep learning techniques with different feature selection approaches. Overall, our study contributes to the growing body of research on the use of machine learning in social media for detecting suicidal ideation and ultimately preventing self-harm and suicide.

\bibliography{anthology}

\end{document}